\pdfoutput=1

\documentclass[11pt]{article}

\usepackage[final]{acl}

\usepackage{times}
\usepackage{latexsym}
\usepackage{booktabs, multirow} 
\usepackage{amsfonts,amsmath,amssymb,amsthm}
\usepackage{makecell}

\usepackage{soul}
\usepackage{color}
\usepackage{booktabs}
\usepackage{multirow}
\usepackage{colortbl}

\usepackage[T1]{fontenc}

\usepackage[utf8]{inputenc}

\usepackage{microtype}

\usepackage{inconsolata}

\usepackage{graphicx}

\title{e-Health CSIRO at RRG24: Entropy-Augmented Self-Critical Sequence Training for Radiology Report Generation}

\author{Aaron Nicolson, Jinghui Liu, Jason Dowling, Anthony Nguyen, \& Bevan Koopman \\
  Australian e-Health Research Centre, CSIRO Health and Biosecurity, Brisbane, Australia \\
  \texttt{aaron.nicolson@csiro.au} \\
}

\begin{document}
\maketitle
\begin{abstract}
The Shared Task on Large-Scale Radiology Report Generation (RRG24) aims to expedite the development of assistive systems for interpreting and reporting on chest X-ray (CXR) images. This task challenges participants to develop models that generate the \textit{findings} and \textit{impression} sections of radiology reports from CXRs from a patient's study, using five different datasets. This paper outlines the e-Health CSIRO team's approach, which achieved multiple first-place finishes in RRG24. The core novelty of our approach lies in the addition of entropy regularisation to self-critical sequence training, to maintain a higher entropy in the token distribution. This prevents overfitting to common phrases and ensures a broader exploration of the vocabulary during training, essential for handling the diversity of the radiology reports in the RRG24 datasets. Our model is available on Hugging Face (\url{https://huggingface.co/aehrc/cxrmate-rrg24}).
\end{abstract}

\section{Introduction}

Machine learning holds the potential to significantly enhance diagnostic processes and clinical reporting, particularly within the field of radiology --- a discipline characterised by high volumes of imaging data. Radiologists are often tasked with interpreting and reporting on hundreds of imaging studies daily, a repetitive process that is susceptible to fatigue and error. Automated systems capable of generating radiology reports from chest X-rays (CXRs) could greatly alleviate this burden by ensuring consistency and potentially reducing diagnostic turnaround times.

The Shared Task on Large-Scale Radiology Report Generation (RRG24) challenges participants to develop automated systems for producing textual reports from CXR images, with a particular focus on the findings and impression sections \cite{xu-etal-2024-overview, delbrouck2022vilmedic}. These sections are crucial as they convey the diagnostic interpretation and clinical significance of a patient's study. The challenge provides a means to benchmark the various models  under uniform conditions, offering insights into which approaches are most effective for CXR report generation. Participants were to train and evaluate their submissions on a dataset formed from five different sources, including MIMIC-CXR \cite{johnson2019mimic}, CheXpert \cite{chambon2024chexpert}, PadChest \cite{bustos2020padchest}, BIMCV COVID-19 \cite{vaya2020bimcv}, and Open-i IU X-ray \cite{demner-fushman_preparing_2016}. This dataset consisted of four subsets, including the \textit{training}, \textit{validation}, \textit{public-test}, and \textit{hidden-test}, where the radiology reports were available for all except the hidden-test set. Finally, RRG24 presents participants with unique challenges to overcome, such as handling studies with missing sections and deciding whether to use a single model or separate models for each section.

This paper outlines the approach taken by team e-Health CSIRO in the RRG24 challenge. For this, we developed a multimodal language model that conditions report generation not only on previously generated words (or subwords), but also on the image embeddings of all the CXRs of a patient's study. We utilised a single model to generate both sections and incorporated special tokens to signify the absence of a section during training. These special tokens were also used to guide the model to generate specific sections during testing.

\begin{figure*}
    \centering
    \includegraphics[scale=0.8]{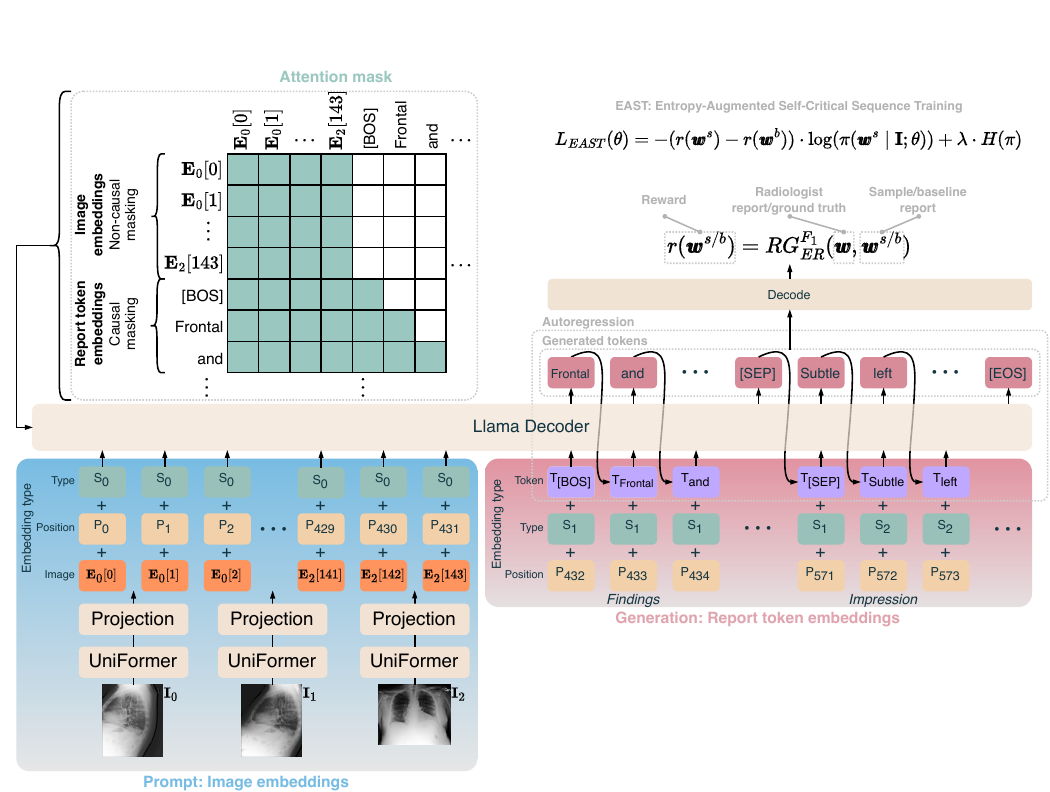}
    \caption{\label{fig:model}e-Health CSIRO's submission into RRG24, named CXRMate-RRG24. \texttt{[BOS]} denotes the \textit{beginning-of-sentence} special token, \texttt{[SEP]} denotes the \textit{separator} special token, and \texttt{[EOS]} denotes the \textit{end-of-sentence} special token. $\textbf{E}_k[i]$ is the $i^{th}$ output of the projected last hidden state of the encoder for the $k^{th}$ image of the study.}
\end{figure*}

A key factor to the performance of our submissions was our modification to the self-critical sequence training (SCST) reinforcement learning (RL) algorithm \cite{rennie_self-critical_2017}. A widely-used technique to enhance RL is to add entropy regularisation into the objective function. This approach boosts exploration and prevents the model from prematurely settling on less optimal actions \cite{mnih_asynchronous_2016}. Hence, we add entropy regularisation to SCST, forming Entropy-Augmented Self-critical sequence Training (EAST). Using EAST, we optimised our model with RadGraph as the reward \cite{delbrouck2022improving}. RadGraph is the primary metric for RRG24; it evaluates the accuracy of a generated report by assessing how well the identified entities and their relationships align with those in a radiologist report. By optimising for this reward, we achieved multiple first-place finishes in RRG24.

\section{Methodology}

\subsection{EAST: Entropy-Augmented Self-critical sequence Training}

Entropy-Augmented Self-critical sequence Training (EAST) builds upon self-critical sequence training (SCST) by incorporating entropy regularisation. This encourages the model to maintain a higher entropy in its token distribution, thereby promoting diversity in token selection and preventing premature convergence on a smaller, selective set of tokens. The loss for SCST is as follows:
\begin{equation}
    L_{SCST}(\theta) = -(r(\pmb w^{s}) - r(\pmb w^{b})) \cdot \log(\pi(\pmb w^{s} \mid\mathbf{I}; \theta)),
\end{equation}
where $r(\pmb w^{s})$ is the reward for the sampled report ($\pmb w^{s} = (w^{s}_1, . . . , w^{s}_M )$ denotes the tokens of length $M$ of the sampled report), $r(\pmb w^{b})$ is the reward for the baseline report ($\pmb w^{b} = ({w}^{b}_1, . . . , {w}^{b}_N )$ denotes the tokens of length $N$ of the baseline report, where the baseline is generated with greedy search), $\mathbf{I} = [I_1, I_2, \ldots, I_K]$ denotes the images of a study (where $K$ is the number of images in the study), $\theta$ represents the parameters of the model, and $\pi(\pmb w^{s} \mid\mathbf{I}; \theta)$ denotes the policy under which $\pmb w^{s}$ is sampled from. As illustrated in Figure \ref{fig:model}, we utilise the RadGraph ER F1-score as the reward \cite{delbrouck2022improving}, where the generated report is either the sample or baseline report, both of which are compared to the radiologist report.

\begin{table*}
    \centering
    \small
    \caption{\label{tab:public}\textbf{Public test set} scores for the findings and impression sections (\textbf{presented as findings/impression}). The order of the leaderboard for RRG24 was determined by RadGraph-F1. The best scores are indicated in boldface.}
    \begin{tabular}{ccccc>{\columncolor[gray]{0.8}}c}
        \toprule
         Team/Method & BLEU-4 & ROUGE-L & BERTScore  & CheXbert-F1  & RadGraph-F1 \\
         \midrule
         \multicolumn{6}{c}{\cellcolor[HTML]{F0F0F0}\textit{e-Health CSIRO}} \\
         EAST & 12.00/\textbf{9.43} & 26.51/26.58 & 54.64/47.81 & 59.18/\textbf{57.73} & 29.46/\textbf{27.01} \\
         SCST & 10.70/8.51 & 26.54/26.30 & 54.79/48.25 & 56.42/55.00 & 27.66/25.04 \\
         TF & 11.63/7.52 & 25.92/23.34 & 51.34/41.46 & 50.73/47.27 & 23.12/20.08 \\
         
         \midrule
         \multicolumn{6}{c}{\cellcolor[HTML]{F0F0F0}\textit{Top three teams besides ours}} \\
         \texttt{tartan} & \textbf{21.59}/- & \textbf{42.03}/- & \textbf{64.34}/- & \textbf{59.70}/- & \textbf{38.05}/- \\
         \texttt{maira} & 12.26/8.68	& 28.00/\textbf{28.40}	& 55.76/\textbf{50.48} & 59.71/56.46	& 26.33/25.89 \\
         \texttt{airi} & 10.13/7.10 & 26.54/25.92 & 53.84/47.18 & 55.49/51.33 & 25.82/24.07 \\
         \bottomrule
    \end{tabular}    
\end{table*}

EAST is formed by adding an entropy term to $L_{SCST}(\theta)$:
\begin{equation}
    L_{EAST}(\theta) = L_{SCST}(\theta) + \lambda \cdot H(\pi)
\end{equation}
where $\lambda$ is a coefficient that determines the weight of the entropy term in the loss function. The entropy is as follows:
\begin{equation}
H(\pi) = -\sum_{v \in \mathcal{V}} \pi(v \mid x; \theta) \log \pi(v \mid x; \theta),
\end{equation}
where $x$ represents the current state (as determined by the image embeddings and the previously generated tokens) and $v$ represents a token from the vocabulary $\mathcal{V}$. This discourages the policy from converging too quickly to deterministic actions, thus encouraging the exploration of a wider set of generated reports.

\subsection{Special Tokens and Missing Sections}

As illustrated in Figure \ref{fig:model}, our model generates both sections. To delineate these sections within the generated text, we utilise a separator token, following CXRMate \cite{nicolson_longitudinal_2024}.\footnote{\url{https://huggingface.co/aehrc/cxrmate}} To accommodate reports during training that have a missing section, we employ two special tokens: \texttt{[NF]} for `no findings' section and \texttt{[NI]} for `no impression' section. They are used in place of the missing sections. They also facilitate the generation of specific sections as needed. For example, if only the impression section is to be generated, \texttt{[BOS][NF][SEP]} can be fed to the decoder to signal that the findings section is not to be generated. Furthermore, to encourage the generation of the impression section, the probability of the \texttt{[NI]} token can be set to zero.

\subsection{Model}

Our model, CXRMate-RRG24, is an evolution of our previous model, CXRMate, and is illustrated in Figure \ref{fig:model}. We utilised UniFormer as the encoder (in particular, the $384 \times 384$ base model warm started with its token labelling fine-tuned checkpoint) \cite{li_uniformer_2023}, which, in preliminary testing, performed comparably to the convolutional vision Transformer (CvT) (which we found to be the best performing encoder for CXR report generation in our previous work \cite{nicolson_improving_2023}) but significantly reduced the training time. The image embedding prompt is formed by processing each image in the study separately with the encoder and then projecting the encoder's last hidden state to match the decoder's hidden size using a learnable weight matrix. Each image was resized using bilinear interpolation so that its smallest side had a length of 384 and its largest side maintained the aspect ratio. Next, the resized image was cropped to a size of $\mathbb{R}^{3 \times 384 \times 384}$. The crop location was random during training and centred during testing. Following \cite{elgendi_effectiveness_2021}, the image was rotated around its centre during training, where the angle of rotation was sampled from $\mathcal{U}{[{-5^{\circ}, 5^{\circ}}]}$. Finally, the image was standardised using the statistics provided with the UniFormer checkpoint. A maximum of five images per study were used during training. If more were available, five were randomly sampled uniformly without replacement from the study.

\begin{table*}
    \centering
    \small
    \caption{\label{tab:hidden}\textbf{Hidden test set} scores for the findings and impression sections (\textbf{presented as findings/impression}). The order of the leaderboard for RRG24 was determined by RadGraph-F1. The best scores are indicated in boldface.}
    \begin{tabular}{ccccc>{\columncolor[gray]{0.8}}c}
        \toprule
         Team/Method & BLEU-4 & ROUGE-L & BERTScore  & CheXbert-F1  & RadGraph-F1 \\
         \midrule
         \multicolumn{6}{c}{\cellcolor[HTML]{F0F0F0}\textit{e-Health CSIRO}} \\
         EAST & \textbf{11.68}/\textbf{12.33} & 26.16/28.32 & 53.80/50.94 & 57.49/\textbf{56.97} & \textbf{28.67}/\textbf{27.83} \\
         SCST & 10.25/10.95 & 26.10/27.34 & 53.88/50.07 & 55.78/54.79 & 27.29/24.97 \\
         TF & 11.12/9.89 & 25.43/24.94 & 51.10/42.49 & 50.02/47.24 & 22.99/21.27 \\
         
         \midrule
         \multicolumn{6}{c}{\cellcolor[HTML]{F0F0F0}\textit{Top three teams besides ours}} \\
         \texttt{maira} & 11.24/11.66 & \textbf{26.58}/\textbf{28.48} & \textbf{54.22}/\textbf{51.62} & \textbf{57.87}/53.27 & 25.48/25.26 \\
         \texttt{airi} & 9.97/10.91 & 25.82/27.46 & 52.42/49.55 & 54.25/52.32 & 25.29/24.67 \\
         \texttt{gla-ai4biomedic} & 7.65/9.60 & 24.35/25.27 & 52.69/48.60 & 46.21/46.74 & 24.13/22.10 \\
         \bottomrule
    \end{tabular}    
\end{table*}

For the decoder, we employed the Llama architecture, which is notable for features such as its rotary positional encoding (RoPE), root mean square normalisation (RMSNorm), and SwiGLU activation function \cite{touvron_llama_2023}. The decoder was initialised randomly and used the CXRMate vocabulary, which was derived from the MIMIC-CXR training set. The hyperparameters of the Llama decoder mirror that of the CXRMate decoder, with six hidden layers, a hidden size of 768, 12 attention heads per layer, and an intermediate size of $3\,072$. Following CXRMate, we added source type embeddings to the input of the decoder to differentiate between findings and impression section tokens, as well as image embeddings. The max number of position embeddings was set to 2048 to accommodate both the image embeddings and the report token embeddings. The maximum number of tokens that could be generated was set to 512, which was also the limit for the radiologist reports during training. During testing, a beam size of four was utilised. Another factor that led to the use of the Llama decoder was the ease of providing a custom attention mask to current implementations.\footnote{https://huggingface.co/blog/poedator/4d-masks} This enabled non-causal masking to be utilised for the prompt and causal masking for the report token embeddings, as shown in Figure \ref{fig:model}. This ensured that the self-attention heads were able to attend to all of the image embeddings at each position.

\subsection{Training}

Two stages of training were performed; teacher forcing (TF) \cite{williams_learning_1989}, followed by RL (either EAST or SCST). \textit{AdamW}~\citep{loshchilov_decoupled_2022} was used for mini-batch gradient descent optimisation with an initial learning rate of 5e-5 for TF and 5e-6 for RL, a mini-batch size of 16 for TF and 8 for RL, a maximum of 32 epochs for TF and 1 epoch for RL, executed on a 94GB NVIDIA H100 GPU with FP32. For RL, validation was performed every $\frac{1}{50}$ of an epoch. The validation macro-averaged CheXbert F1 was the monitored metric for checkpoint selection. For RL, the sample report was generated with top-\textit{k} sampling ($k=50$). During RL, the encoder was frozen. For EAST, the entropy weight ($\lambda$) was set to 0.05.

\section{Results and Discussion}

The results for our key submissions on the public and hidden test sets are shown in Tables \ref{tab:public} and \ref{tab:hidden}, respectively. The metrics utilised for RRG24 include BLEU-4 \cite{papineni2002bleu}, ROUGE-L \cite{linAutomaticEvaluationMachine2004}, BERTScore \cite{zhang2019bertscore}, CheXbert-F1 \cite{smit2020combining}, and RadGraph-F1 \cite{delbrouck2022improving}, the later of which is the primary metric used to rank the teams. Here, we compare TF, to SCST, and to our proposed method, EAST. EAST attained a higher score than TF for each metric, something SCST was not able to do (TF attained a higher BLEU-4 score than SCST for the findings section of both test datasets). 

Comparing EAST to SCST, SCST attained a higher ROUGE-L score on the public-test findings sections, and a higher BERTScore on the public-test findings and impression sections, as well as the hidden-test findings sections. For all other cases, EAST demonstrated an improvement over SCST. Policies trained with entropy regularisation often have improved generalisation, as they have learnt to consider a broader set of possible actions. This may have led EAST to be more robust to the differing characteristics of each of the datasets used in the public and hidden test sets. With EAST, team e-Health CSIRO achieved a first-place finish amongst participants for the public-test impression sections and the hidden-test findings and impression sections. We also also achieved a second-place finish for the public-test findings sections. For a comparison of CXRMate-RRG24 to state-of-the-art methods in the literature, please see \citet{nicolson_impact_2024}.

\subsection{Conclusion}

Our proposed approach, EAST, was able to generate reports that were quantitatively more aligned with radiologist reports than those generated using SCST. By incorporating entropy regularisation, EAST is able to maintain a higher diversity in token selection and mitigate overfitting to maintain generalisability. This was likely crucial in handling the varied characteristics of the datasets used in RRG24. While EAST shows promise, a more thorough investigation is required to validate its potential, including the impact of varying the entropy coefficient.



\bibliography{bibliography}




\end{document}